# Parcellation of fMRI datasets with ICA and PLS - a data driven approach

Yongnan Ji[1], Pierre-Yves Hervé[2], Uwe Aickelin[1], and Alain Pitiot[2]

[1] School of Computer Science, University of Nottingham,
Jubilee Campus, Wollaton Road, Nottingham, NG8 1BB, U.K.
yxj@cs.nott.ac.uk,
[2] Brain and Body Centre, University of Nottingham,
University Park, Nottingham, NG7 2RD, U.K.

**Abstract.** Inter-subject parcellation of functional Magnetic Resonance Imaging (fMRI) data based on a standard General Linear Model (GLM) and spectral clustering was recently proposed as a means to alleviate the issues associated with spatial normalization in fMRI. However, for all its appeal, a GLM-based parcellation approach introduces its own biases, in the form of a priori knowledge about the shape of Hemodynamic Response Function (HRF) and task-related signal changes, or about the subject behaviour during the task.
In this paper, we introduce a data-driven version of the spectral clustering parcellation, based on Independent Component Analysis (ICA) and Partial Least Squares (PLS) instead of the GLM. First, a number of independent components are automatically selected. Seed voxels are then obtained from the associated ICA maps and we compute the PLS latent variables between the fMRI signal of the seed voxels (which covers regional variations of the HRF) and the principal components of the signal across all voxels. Finally, we parcellate all subjects data with a spectral clustering of the PLS latent variables.
We present results of the application of the proposed method on both single-subject and multi-subject fMRI datasets. Preliminary experimental results, evaluated with intra-parcel variance of GLM t-values and PLS derived t-values, indicate that this data-driven approach offers improvement in terms of parcellation accuracy over GLM based techniques.

## 1 Introduction

Inter-subject parcellation based on a standard General Linear Model (GLM) and spectral clustering was recently proposed as a means to alleviate the issues associated with spatial normalization in the analysis functional Magnetic Resonance Imaging (fMRI) datasets: lack of true anatomical correspondence, inaccuracy of the normalization process (see [1] for an in-depth overview), etc. In a parcellation framework, voxels are first clustered into functionally homogeneous regions or parcels. Then, the parcellations are then homogenised across subjects, so that statistics can be carried out at the parcel level rather than at the voxel level. Here we focus on the optimization of the first step of the parcellation scheme.

We present a data-driven, model-free, parcellation technique, based on Independent Component Analysis (ICA) and Partial Least Squares (PLS, [2]) instead of a GLM, so as to use more of the information contained within the fMRI time series. First, a number of independent components are automatically selected. Seed voxels are then obtained from the associated ICA maps and we compute the PLS latent variables between the fMRI signal of the seed voxels (which covers regional variations of the stimuli related BOLD responses) and the principal components of the signal across all voxels. Finally, we parcellate all subjects data with a spectral clustering of the PLS latent variables. We also introduce PLS t-values as an alternative way to validate parcellation results.

We detail our approach in the following Section 2. Preliminary results are given in Section 3, where we also compare them to GLM-based parcellation.

## 2 Materials and Methods

### 2.1 Datasets and Preprocessing

We applied our method to two functional datasets: a single-subject fMRI experiment with a standard finger tapping task and a multi-subject experiment where volunteers were presented with hand gestures or face expressions [3].

Single-subject data were acquired on a Philips Intera 1.5T with a TR of 3s and a sequential finger tapping task auditorily paced with a metronome. The auditory signals were given every 0.6 seconds. The digit order of the tapping was 1 - 3 - 2 - 4, repeated 6 times in each period, with a 14.4 second rest between periods. The period of one on-and-off block was then 28.8 seconds.

Multi-subject data, our main concern in this paper, were acquired from 25 subjects viewing angry gestures or expressions. Scanning was performed on a Philips Intera 1.5T, with TR=3s. During the scan, four types of visual stimuli are given to the subjects, which are angry hand gestures, neural hand gestures, angry facial expression and neural facial expression.

Both datasets were preprocessed with FSL for slice-timing, motion correction and registration [4, 5].

### 2.2 Independent Components Selection

For the multi-subject experiment, we used FSL to decompose the input fMRI data into independent components (ICs). We obtained between 30 and 60 ICs per subject, for a total of 1203 ICs. Here we propose to use a hierarchical clustering approach, similar in spirit to Partner Matching [6] as a means to find the ICs that best capture the Blood Oxygen Level Dependent Haemodynamic (BOLD) response to the stimuli. This method is based on the assumption that very few of the 1203 ICs contain information about the stimuli-related BOLD responses. Consequently, the task-related ICs should be more similar with each other than with the other ICs since they share the same source. We aim to group those ICs that correspond to the response to the same task features in different subjects together into one cluster. The other ICs which do not contain relevant (i.e. task-related) information should be grouped inside another cluster.

We take those constraints into account when design the similarity function to be used in our hierarchical clustering.

Let $N_a$ and $N_b$ be the number of ICs for subjects A and B respectively, with $\mathbf{IC}_i^A$ and $\mathbf{IC}_j^B$ the $i$th IC of subject A and $j$th IC of subject B. $t = 1, \ldots, T$ is the time index. Their correlation coefficients is given by:

$$\rho(\mathbf{IC}_i^A, \mathbf{IC}_j^B) = \frac{\sum_{t=1}^{T}(\mathbf{IC}_i^A(t) - \overline{\mathbf{IC}_i^A})(\mathbf{IC}_j^B(t) - \overline{\mathbf{IC}_j^B})}{\sqrt{\sum_{t=1}^{T}(\mathbf{IC}_i^A(t) - \overline{\mathbf{IC}_i^A})^2}\sqrt{\sum_{t=1}^{T}(\mathbf{IC}_j^B(t) - \overline{\mathbf{IC}_j^B})^2}}, \quad (1)$$

The normalized correlation coefficients $\rho_{\text{norm}}$ is:

$$\rho_{\text{norm}}(\mathbf{IC}_i^A, \mathbf{IC}_j^B) = \frac{\rho(\mathbf{IC}_i^A, \mathbf{IC}_j^B) - \text{mean}(\rho(\mathbf{IC}_i^A, \mathbf{IC}_j^B)|_{j=1,2,\ldots,N_b})}{\text{std}(\rho(\mathbf{IC}_i^A, \mathbf{IC}_j^B)|_{j=1,2,\ldots,N_b})} \quad (2)$$

Since the aim of the clustering is to put similar ICs from different subjects into one cluster, all the ICs of the same cluster should come from different subjects, therefore we need to set the similarity between ICs of the same subject to 0. The similarity between two ICs is finally defined as

$$S(\mathbf{IC}_i^A, \mathbf{IC}_j^B) = \begin{cases} 0 & \text{if } A = B \\ \min(\rho_{\text{norm}}(\mathbf{IC}_i^A, \mathbf{IC}_j^B), \rho_{\text{norm}}(\mathbf{IC}_j^B, \mathbf{IC}_i^A)) & \text{other wise} \end{cases} \quad (3)$$

In the case of the single-subject data, the ICs representing BOLD signals could not be selected by comparing ICs across subjects as above. Therefore, we manually picked those ICs that best matches the canonical HRF-convoluted task design from the 34 ICs produced by FSL.

### 2.3 Seed Selection

In order to calculate the PLS latent variables that best capture the BOLD response, a number of seeds representing different active regions should be selected. For instance, in a GLM-based parcellation approach, we could select as seeds the voxels with the largest t-values. Here, we pick them on the basis of the ICA results.

Within each IC map [5], the first seed is chosen as the voxel with the largest value. The second seed is then chosen, amongst the voxels at least R voxels away from the first seed voxel, as the voxel with the largest IC map value. The iterative process is repeated until all the seeds have been selected.

In the multi-subject case, two IC maps were used. We picked R = 6 voxels and obtained $N_{\text{seed}} = 15$ seeds for each map. In the single-subject data, 30 seeds were selected from each IC map with R = 6.

## 2.4 PCA/PLS Feature Space for Parcellation

Let $X_{V \times T} = x_{ij}$ denote the data matrix, where each row corresponds to the fMRI signal of a given voxel. Then, we propose to denoise the signal with PCA. We first center the signal at each voxel by substracting its mean, before decomposing $X_{V \times T}$ into $P_{PCA}$ and $T_{PCA}$ using Principal Component Analysis (PCA) as $X_{V \times T} = P_{PCA} T'_{PCA}$ [7, 8]. $T'_{PCA}$ is the transpose of the PCA score matrix of X (the matrix whose columns are the Principal Components (PCs) of the fMRI data), and $P_{PCA}$ is the PCA loading matrix. Ranking the PCs according to the variance they cover, the first few PCs usually have exceptionally high variances. And the last PCs are slow-variant artefacts. These PCs are considered as noise and removed.

Let $D_{T \times N_{seed}}$ represent the fMRI signals of the seed voxels, where each column correspond to the fMRI signal in a given seed. We then use the PCs in matrix $T_{PCA}$ for the prediction of D with Partial Least Square (PLS). The original design of PLS is to predict D with the components decomposed from $T_{PCA}$ and D as regressor. These components, the latent variables, should contain the information from both $T_{PCA}$ and D. Here PLS is used to calculate the time series components that represent the individual specific functional activity signals. We decompose $T_{PCA}$ into the product of $T_{PLS}$ and $P'_{PLS}$ with $T'_{PLS} T_{PLS} = I$. D is predicted as $\hat{D} = T_{PLS} B C'$, where the columns of $T_{PLS}$, $t_i, i = 1, 2, ..., K$, are the latent vectors of size $T \times 1$. B is a diagonal matrix with the "regression weights" as diagonal elements and C is the "weight matrix" of the dependent variables [9].

Given $T_{PCA}$ and D, the latent vectors could be chosen in a lot of different ways. The canonical way is to find the latent vectors that maximize the covariance between the columns of $T_{PLS}$ and D. Specifically, the first latent vector is calculated as $t_1 = T_{PCA} w_1$ and $u_1 = D c_1$ with the constrains that $t'_1 t_1 = 1$, $w'_1 w_1 = 1$ and $t'_1 u_1$ be maximal. Then first component is subtracted from $T_{PCA}$ and D, and the rest latent variables are calculated iteratively as the above until $T_{PCA}$ becomes a null matrix. The first PLS latent variables are the signals of interest. Let $X_0$ be derived from X after the signal variance has been revmoed: $x_0 = x/||x||$, where x and $x_0$ are the row vectors of X and $X_0$. We use the covariances between fMRI signals and latent variables, $r_i = X_0 t_i$, as feature space for parcellation.

## 2.5 Parcellation method

Following Thirion et al.[1], we chose spectral clustering for parcellation. This method represents the relationships between voxels as a graph whose vertices correspond to the voxels with the functional distance between voxels (GLM-based in their approach) associated to the edges. The complete distance matrix $\Delta_G$ between all pairs of voxels is obtained by integrating the local distances along the paths in the graph. Singular Value Decomposition (SVD) is applied to the centered square distance matrix $\Delta_G$. Finally, they apply C-means clustering to the singular vectors with largest singular values. Please see [1] for details.

Here, rather than using GLM to define local distances, we use our $r_i$'s:

$$d(v,w) = ||\mathbf{r}(v) - \mathbf{r}(w)|| = \sqrt{(\mathbf{r}(v) - \mathbf{r}(w))(\mathbf{r}(v) - \mathbf{r}(w))'}, \quad (4)$$

where v and w are two neighbouring voxels and $\mathbf{r}(v) = [r_1(v)\ r_2(v))...\ r_K(v)]$.

## 3 Results

### 3.1 Intra-parcel functional homogeneity

As in [1], we use the intra-parcel functional variance to assess the quality of the parcellation. However, instead of using GLM parameters to represent the functional information, we use GLM t-values and PLS t-values (described below) for each regressor as functional features. Let $N_r$ be the number of regressors and $\mathbf{f}^i \in R^{N_r \times 1}$ be the vector of t-values for voxel i. For any parcel p, the functional variance of p, v(p), is:

$$v(p) = \sqrt{\sum_{k=1}^{N_r}(\text{std}(f_k^i))^2}, \quad \text{where, } \mathbf{f}^i = [f_1^i\ f_2^i\ \ldots\ f_{N_r}^i]' \text{ and } i \in p \quad (5)$$

The distribution of v(p) across all parcels is used to compare the accuracy of the parcellations.

**PLS t-values** Given a design matrix $Y \in R^{T \times N_r}$, where $y_k \in R^{T \times 1}$ is the kth column of Y, instead of using D, the regressor $y_k$ is used to calculate latent variables as in section 2.4. If $r_k$ is the covariance between the fMRI time series and the first latent variable, then,

$$t = \frac{r\sqrt{T-2}}{1-r^2} \quad (6)$$

has a t-distribution with $T-2$ degrees of freedom. The null hypothesis of this test is that the signal of that voxel is not covariant with the PLS components. Thus, we can generate statistical maps to represent the significance of the covariance between the signals in each voxel and the first latent variable calculated from data and the kth regressor of the design matrix.

### 3.2 Results on single-subject data

As mentioned above in section 2.2, our automatic IC selection approach cannot be applied to single subject data. Here, we manually selected the IC whose time course best matches the experiment design, to be used in the seed selection process. The fMRI signals of the whole brain are decomposed into PCs. It should be noted that the PCs covering the largest variance in X are not the most interesting signals in the fMRI dataset. Indeed it appears that the respiratory, cardiac or instrumental artifacts may have a larger influence on the BOLD signal than the task. Such noise-related PCs are removed before the PLS step.

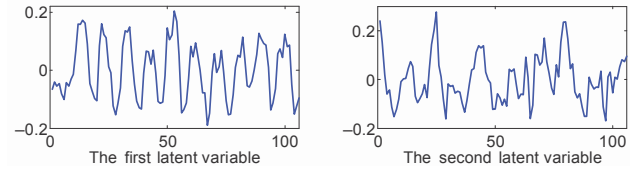

Fig. 1. First two PLS components

In Fig. 1, the first two latent variables are shown. The shape of the first latent variable matches the experiment design. It can be considered as a subject-specific response model, which will allow a better detection of task-related activity. Based on the latent variables, the whole brain is parcellated into 600 parcels using a spectral clustering as explained in section 2.5. The intra-parcel variances of GLM t-values and PLS t-values are used to compare parcellation results based on GLM and PLS. Here, the bars illustrate the mean, the first and the third quartile of the t-values variance of 600 parcels from each method. From Fig. 2, we can see that with both functional measures, spectral clustering with PLS increases the intra-parcel functional homogeneity. One latent variable is optimal for the parcellation of this dataset.

### 3.3 Results on multi-subject data

Using the similarity matrix described in section 2.3 and Ward's linkage, we grouped all the ICs into three clusters. The ICs in cluster 1 match the first and second task regressors. The ICs in cluster 2 match the third and fourth task regressor. Meanwhile, in feature space, these two clusters keep large distances from the rest of the ICs. There are 20 ICs in cluster 1 from 19 subjects and 20 ICs in cluster 2 from 20 subjects. For each subject we use the ICs from these for sampling seed voxels. If a subject doesn't have an IC in cluster 1 or 2, we use the ICs that are closest to those clusters to sample the seeds.

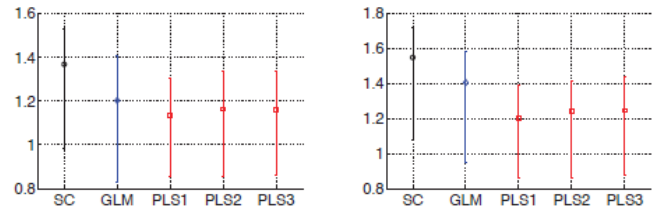

Fig. 2. Comparison of Parcellation results from Spatial Clustering (SC), Spectral Clustering with GLM (GLM), Spectral Clustering with 1 PLS latent variable (PLS1), 2 PLS latent variable (PLS2) and 3 PLS latent variable (PLS3)

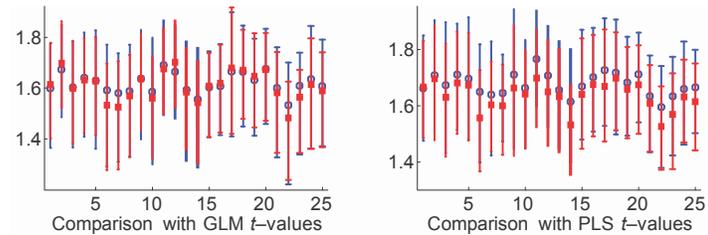

Fig. 3. Comparison of functional intra-parcel homogeneity

All the datasets are then parcellated into 600 parcels. The functional variances of GLM parcellation and ICA-PLS parcellation on 25 subjects are compared in Fig. 3. Subjects are split across the horizontal axis. The parcels respond to the stimuli of angry hand gestures are shown in Fig. 4. The activation and intra-parcel functional variance are evaluated with GLM t-values. Similarly, in Fig. 5, the parcellation results are evaluated with PLS t-values.

## 4 Conclusion and Future Work

We presented a data driven method for parcellation of fMRI data. Preliminary experimental results indicate that such approach adapts to the variability of the BOLD response across subjects and increase the accuracy of the parcellation. The cost of this improvement is the complexity of parcellation. Future work will tackle the homogenisation of those parcellations across different subjects.

Acknowledgement This research is funded by the European Commission FP6 Marie Curie Action Programme (MEST-CT-2005-021170).

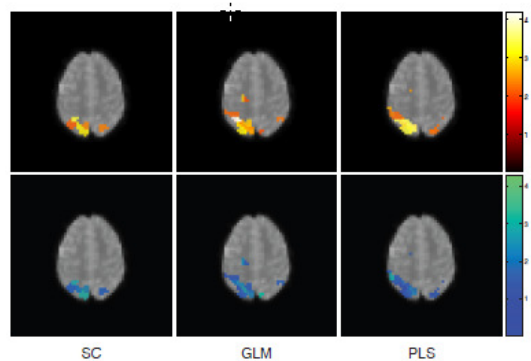

Fig. 4. Parcels respond to stimuli. Parcels with average GLM t-values larger than 2 are shown. The first row shows activation. The second row shows intra-parcel variance. Three columns show the results from three parcellation methods.

88

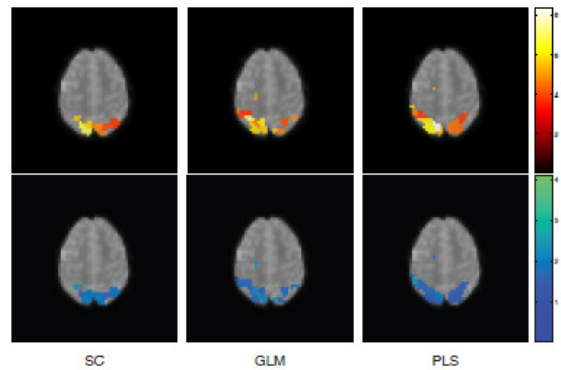

Fig. 5. Parcels respond to stimuli. Parcels with average PLS t-values larger than 3 are shown.